%% file: example.tex
\documentclass{article}

\usepackage[preprint]{corl_2026} 
\usepackage[utf8]{inputenc} 
\usepackage[T1]{fontenc}    
\usepackage{hyperref}       
\usepackage{url}            
\usepackage{booktabs}       
\usepackage{amsfonts}       
\usepackage{nicefrac}       
\usepackage{microtype}      
\usepackage{xcolor}         
\usepackage{pifont}

\usepackage{makecell}

\usepackage{multicol}
\usepackage{subcaption}
\usepackage{mathrsfs}
\usepackage{amsfonts}
\usepackage{colortbl}               
\usepackage{amsfonts}       
\usepackage{nicefrac}       
\usepackage{microtype}  
\usepackage{marvosym}
\usepackage{array} 
\usepackage{graphicx}
\usepackage{amsmath}
\usepackage{amssymb}
\usepackage{multirow}
\setcitestyle{numbers,square}
\usepackage{algorithm}  
\usepackage{algorithmicx} 
\usepackage{algpseudocode}
\usepackage{caption}
\usepackage{subfloat}
\usepackage{tabularx} 
\usepackage{ragged2e} 
\usepackage{makecell}
\usepackage{amsfonts}
\usepackage{xspace}
\usepackage{anyfontsize}
\usepackage{nicefrac}  
\usepackage[accsupp]{axessibility} 
\usepackage{float}
\usepackage{cleveref}
\usepackage{wrapfig}

\title{DeCAL: Towards Physically-Grounded Dexterous Vision-Language-Action  Models via Contact-Aware Latent Co-Imagination}

\author{
        Yankai Fu\textsuperscript{1,2}$^{*}$, 
        Ning Chen\textsuperscript{1,2}$^{*}$, 
        Junkai Zhao\textsuperscript{2}$^{\dagger}$, 
        Heng Zhang\textsuperscript{1}, \\
        \textbf{Guocai Yao\textsuperscript{2}, 
        Pengwei Wang\textsuperscript{2}, 
        Zhongyuan Wang\textsuperscript{2}, 
        Shanghang Zhang\textsuperscript{1,2}\textsuperscript{\Letter} \vspace{0.3em}} \\
    \textsuperscript{1}State Key Laboratory of Multimedia Information Processing, School of Computer Science, \\
    Peking University; \textsuperscript{2}Beijing Academy of Artificial Intelligence \\
    $^{*}$Equal contribution, $^{\dagger}$Project leader, \textsuperscript{\Letter}Corresponding author
    \vspace{0.3em}\\
    \textbf{Project Webpage:} \href{https://aureleopku.github.io/DeCAL}{https://aureleopku.github.io/DeCAL}
    \vspace{-0.5em}
}

\begin{document}
\maketitle


\begin{center}
    \vspace{-1em}
    \captionsetup{type=figure}
    \includegraphics[width=\textwidth]{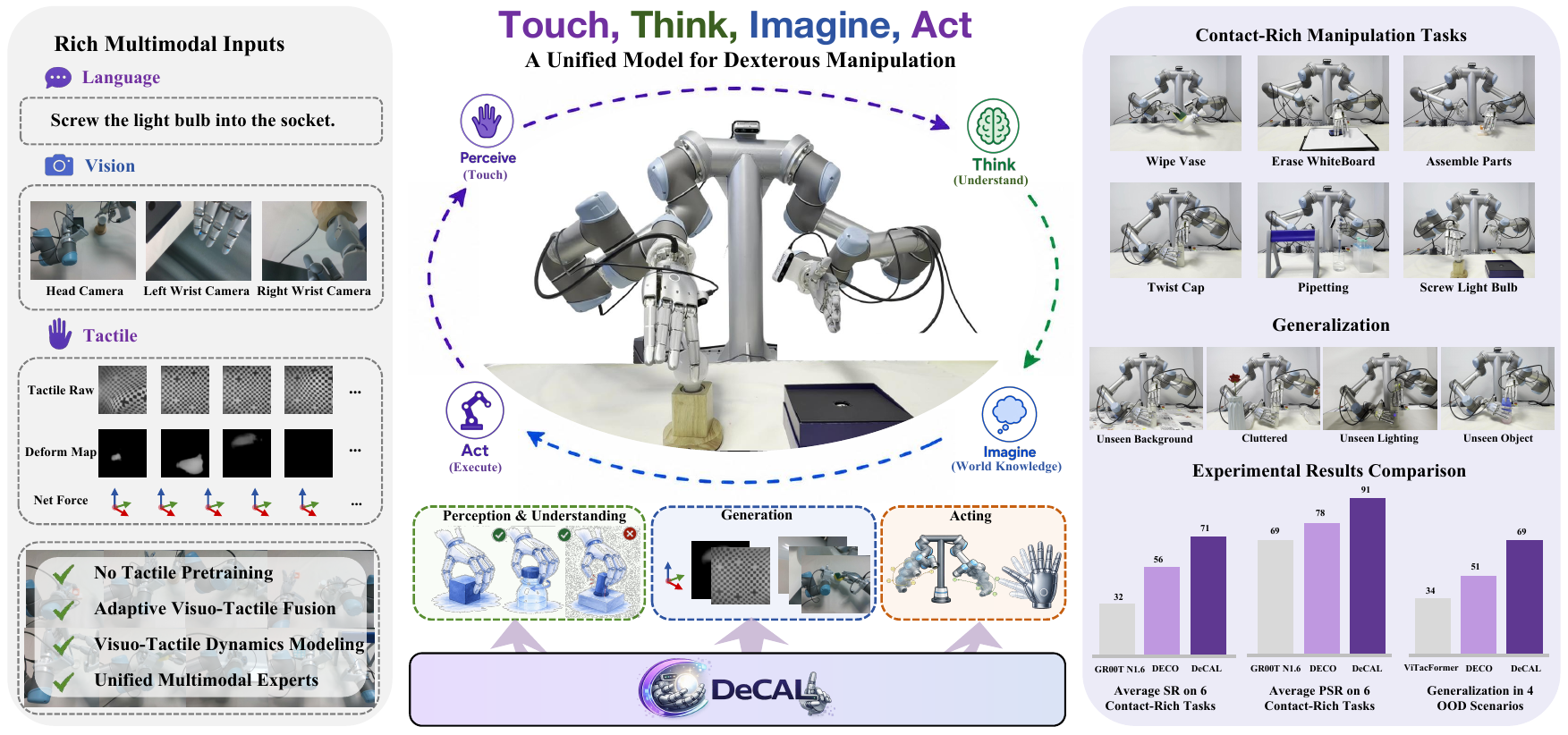}
    \captionof{figure}{We present DeCAL, a physically-grounded dexterous VLA model that unifies perception, understanding, imagination, and action within a framework. Equipped with rich multimodal inputs, DeCAL achieves strong performance across diverse contact-rich dexterous manipulation tasks and demonstrates robust generalization to unseen scenarios.}
\end{center}

\begin{abstract}
Dexterous manipulation involves contact-rich and fine-grained interactions with the physical world, posing significant challenges for existing vision-language-action (VLA) models due to severe visual occlusions and complex contact dynamics. 
While recent works have incorporated tactile sensing into robotic manipulation, most approaches still rely on homogeneous multimodal fusion, lacking adaptive tactile integration and explicit modeling of physical dynamics.
In this work, we present DeCAL, a physically-grounded dexterous vision-language-action model that unifies understanding, imagination and action generation for contact-rich dexterous manipulation. Built upon a Mixture-of-Transformers (MoT) architecture, DeCAL leverages specialized experts for each capability while enabling efficient information flow among them. To effectively leverage tactile information, we introduce Adaptive Visuo-Tactile Fusion that dynamically regulates tactile interactions via a contact-aware gating strategy. Furthermore, we propose Visuo-Tactile Latent Co-Imagination to jointly model visual and tactile dynamics, equipping the policy with implicit physical world knowledge. Experimental results show that DeCAL consistently achieves state-of-the-art performance across all tasks, attaining a 71\% average success rate and an 83.4\% progress success rate, while also demonstrating strong generalization to unseen scenarios.

\end{abstract}



\section{Introduction}
Dexterous manipulation plays a fundamental role in human daily life \cite{chen2022towards, fu2025cordvip}, enabling fine-grained and contact-rich interactions with the physical world. Motivated by recent advances in vision-language models \cite{bai2025qwen3, team2025robobrain, team2025gemini, yang2025magma}, vision-language-action (VLA) models have achieved remarkable progress in robotic manipulation \cite{intelligence2025pi_, kim2025fine, team2024octo, zitkovich2023rt}, exhibiting strong capabilities in visual perception and semantic understanding. Building upon large-scale egocentric human demonstrations~\cite{hoque2025egodex, grauman2024ego, punamiya2026egoverse}, recent works~\cite{yang2025egovla, luo2026being, cai2025n, fu2025metis} have extended VLA models to dexterous manipulation and achieved promising results on multi-fingered manipulation tasks. However, these approaches still struggle with contact-rich and fine-grained interactions due to severe visual occlusions, complex contact dynamics, and limited physical observability from visual inputs alone.

Tactile feedback~\cite{patel2020digger, lambeta2020digit, tomo2017covering, zhang2022tac3d, ye2026data} provides direct access to physical interaction signals, including contact state, force variation, slip, and object deformation, which is essential for dexterous manipulation~\cite{guzey2024see, yuan2024robot, lin2025learning, liu2024masked, dave2024multimodal}. Some studies~\cite{heng2025vitacformer, li2025adaptive, wu2025canonical} have incorporated tactile sensing into robotic manipulation systems to improve contact awareness and manipulation robustness~\cite{huang2025spatially, li2026deco, he2025foar}. However, most existing approaches primarily treat tactile inputs as auxiliary sensory signals and perform homogeneous multimodal fusion, lacking adaptive tactile interaction modeling. Moreover, existing methods are largely reactive and fail to explicitly model future interactions, limiting their ability to reason about evolving contact transitions during manipulation.

In this work, we investigate these problems by introducing DeCAL, a physically-grounded dexterous vision-language-action model that enables effective utilization of tactile signals for contact-rich manipulation. First, DeCAL is built upon a Mixture-of-Transformers architecture composed of three collaborative experts for understanding, imagination, and action generation. The three experts perform directional knowledge sharing through joint attention, providing implicit guidance for coherent and fine-grained dexterous manipulation. Second, we introduce an adaptive visuo-tactile fusion mechanism that dynamically injects tactile information via a contact-aware gating strategy, allowing the model to selectively emphasize tactile cues during physical interactions. Finally, we propose visuo-tactile latent co-imagination, which jointly models visuo-tactile dynamics through future-oriented latent imagination, equipping the policy with implicit world knowledge.

To comprehensively evaluate DeCAL, we conduct extensive real-world experiments on diverse contact-rich dexterous manipulation tasks. The comparative results demonstrate that DeCAL consistently achieves state-of-the-art performance, outperforming the strongest baseline by more than 15\% average success rate. Meanwhile, DeCAL maintains efficient real-time performance, achieving an average inference latency of 0.27s per action chunk. Furthermore, comprehensive ablation studies verify the effectiveness of each proposed component, highlighting the promise of effectively leveraging tactile information for dexterous manipulation. In summary, our contributions are as follows: 

\begin{itemize}
    \item We present DeCAL, a dexterous Vision-Tactile-Language-Action framework that unifies understanding, generation, and action for effective visuo-tactile representation learning.
    \item We introduce Adaptive Visuo-Tactile Fusion and Visuo-Tactile Latent Co-Imagination to facilitate contact-rich and fine-grained dexterous manipulation.
    \item We demonstrate the effectiveness and generalization of our method through a range of real-world experiments.
\end{itemize}

\section{Related Work}

\subsection{Vision-Language-Action Model}
Vision-Language-Action (VLA) models have emerged as a powerful paradigm for robotic control~\cite{intelligence2025pi_, kim2024openvla, team2024octo, fu2025metis, luo2026being, ye2026self}, leveraging the broad semantic and world knowledge embedded in large-scale pre-trained foundation models~\cite{team2025robobrain, team2025gemini, yang2025magma}. By fine-tuning these models on diverse robotic datasets~\cite{open_x_embodiment_rt_x_2023, wu2024robomind, wu2025robocoin, khazatsky2024droid}, researchers have demonstrated successful cross-domain capability transfer from general intelligence to specific manipulation tasks. Representative examples include $\pi_{0.5}$~\cite{intelligence2025pi_}, and GR00T N1.6~\cite{bjorck2025gr00t}.
More recently,  a growing line of work, including InternVLA-A1~\cite{cai2026internvla}, MoTus~\cite{bi2025motus}, and LAST$_0$~\cite{liu2026last}, has advanced VLA research by adopting Mixture-of-Transformer (MoT) architectures. These approaches seamlessly integrate perception (understanding), generation, and action within a unified framework, achieving strong performance across a variety of complex manipulation tasks. 
Despite these advances, existing VLA models remain predominantly vision-centric and lack explicit modeling of physical contact dynamics, limiting their performance in tactile-intensive scenarios involving severe visual occlusions and fine-grained physical interactions.
To address this, we integrate tactile sensing into VLA world modeling by jointly learning visual evolution and contact dynamics, enabling more fine-grained and physically grounded manipulation.

\subsection{Tactile for Dexterous Manipulation}
Dexterous manipulation poses a fundamental challenge for vision-only policies, as the intricate physical structure of multi-fingered hands often leads to severe self-occlusion and ambiguous contact observations~\cite{fu2025cordvip, zhang2026fingervip}.
To mitigate this, recent studies have explored the integration of haptic feedback to maintain operational continuity under visual impairment~\cite{xue2025reactive, qi2023general, guzey2023dexterity, yin2023rotating, guzey2024see, yuan2024robot, yuan2026ftp}.
For instance, some studies~\cite{wu2025canonical, funabashi2022multi, zheng2026omnivta, yang2024binding} learn rich tactile representations through self-supervised objectives, capturing contact geometry and force-related cues to facilitate efficient downstream policy learning. Another line of work~\cite{lou2026dream, heng2025vitacformer, li2025adaptive} predicts future tactile signals to guide action generation, showing the potential of tactile foresight for improving visuo-tactile coordination and contact-aware control. Despite these advances, how to effectively and adaptively couple visual and tactile signals still remains underexplored~\cite{qi2023general, li2026deco}.
Distinct from these approaches, our work integrates tactile sensing into VLA policy learning, leveraging the model's multimodal understanding of language, vision, and physical interactions~\cite{bai2025towards}. By introducing a unified dynamics modeling framework and an adaptive learning mechanism, our model enables more robust visuo-tactile fusion, allowing the agent to dynamically prioritize visual and tactile cues according to task demands and environmental constraints.

\section{Robot System Setup}
Our system consists of a pair of 6-DoF UR5 robotic arms and two 22-DoF SharpaWave five-fingered dexterous hands. Visual observations are captured from two wrist-mounted cameras and one egocentric camera, all using Intel RealSense D435. Each fingertip is equipped with a high-resolution ($320 \times 240$) vision-based tactile sensor developed by Sharpa, enabling fine-grained perception of contact dynamics during manipulation. Specifically, a built-in camera inside each sensor captures the deformation of the elastic surface during physical interaction. Based on the raw tactile images and visuo-tactile processing algorithms, we represent tactile signals in three complementary forms:

\begin{itemize}
    \item \textbf{Raw Image (R).} Raw tactile observations are directly captured by the built-in camera inside each visuo-tactile sensor, recording the contact patterns and elastic surface deformations.
    \item \textbf{Net Force (F).} Each fingertip outputs a 6-DoF force representation consisting of 3-axis forces and torques. The force signals are estimated from raw tactile observations using a pretrained regression model, enabling force-aware tactile perception.
    \item \textbf{Deform Map (M).} Each fingertip outputs a deformation depth map represented as a 2D image through a translation model~\cite{su2026tacmap}, where each pixel indicates the local deformation depth on the tactile surface, providing fine-grained geometric contact information.
\end{itemize}

\section{Method}

\begin{figure}
    \centering
    \captionsetup{type=figure}
    \includegraphics[width=\textwidth]{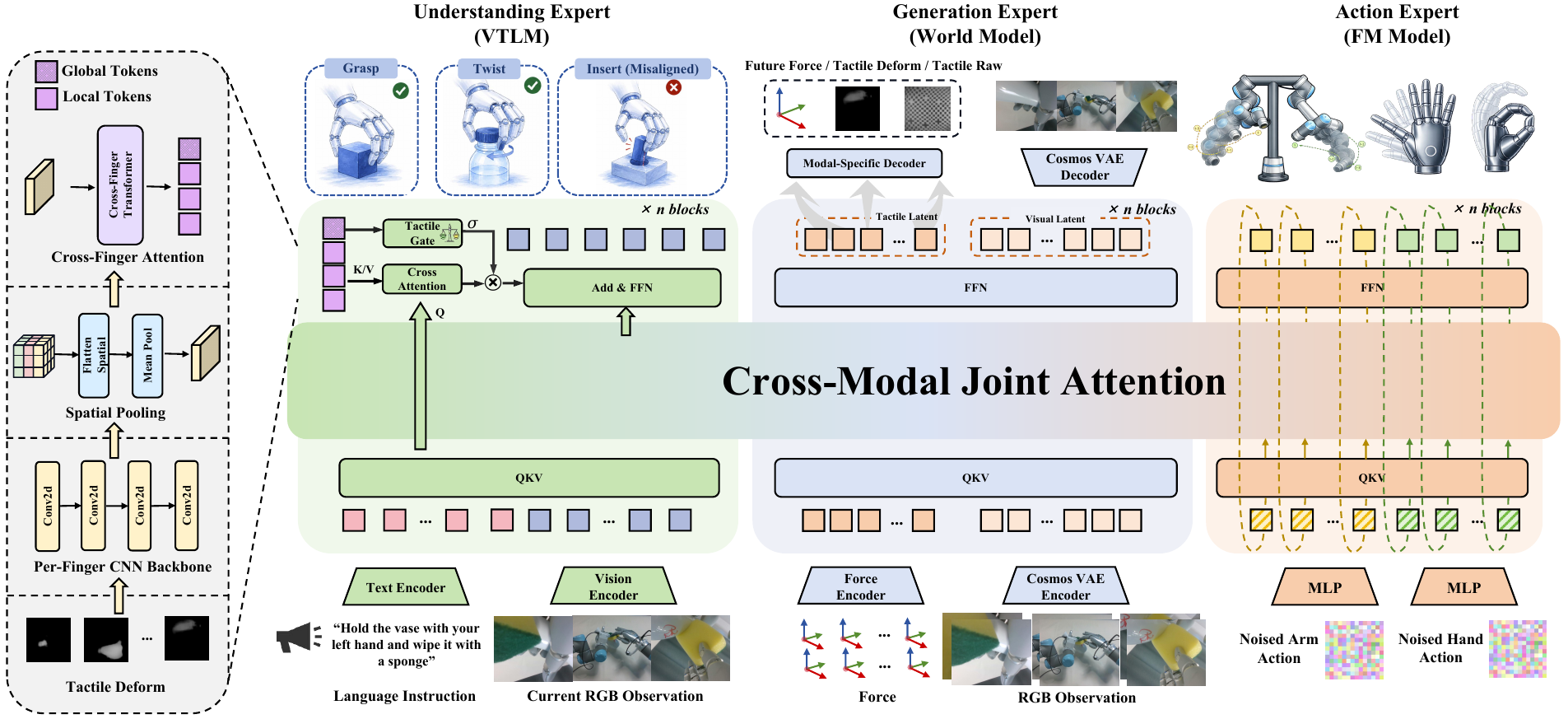}
    \captionof{figure}{DeCAL is built upon a MoT architecture that unifies scene understanding, visuo-tactile dynamics foresight, and action generation. The Action Expert employs Factorized Flow Matching to decouple arm and hand motion, enabling better coordination and dexterous manipulation.}
    \label{fig:pipeline}
    \vspace{-1em}
\end{figure}
\subsection{Preliminaries}
\label{Preliminaries}
The goal of our robot policy is to predict an action chunk from multimodal observations. Most existing VLA models lack explicit modeling of future interaction dynamics, limiting their ability to reason about fine-grained physical interactions. To address this, we propose DeCAL $\pi_\theta$, a unified VLA framework that jointly performs understanding, imagination, and action generation. Given the current observation $o_t = (I_t, H_t, s_t)$, and the language instruction $l$, where $I_t$ denotes the visual observations, $H_t$ denotes the tactile observations, and $s_t$ represents the robot state. DeCAL is trained to jointly maximize the likelihood of future visuo-tactile latent $z_{t+H}$ and future actions $a_{t+1:t+H}$:
\begin{equation}
\max_{\theta}
~
\mathbb{E}_{(o_t, l, a_{t+1:t+H}, z_{t+H}) \sim \mathcal{D}}
\left[
\log \pi_{\theta}(a_{t+1:t+H}, z_{t+H} \mid o_t, l)
\right].
\end{equation}

\subsection{Model Architecture}
\label{Model Architecture}
DeCAL adopts a Mixture-of-Transformers (MOT) architecture that seamlessly integrates scene understanding, visuo-tactile dynamics foresight, and action generation within a unified framework. As illustrated in \Cref{fig:pipeline}, the framework consists of three specialized transformer experts. 

\noindent\textbf{Understanding Expert.}
The understanding expert is built upon Qwen3-VL~\cite{bai2025qwen3} for its strong multimodal understanding capability. 
Given language instructions and multi-view visual observations, the inputs are first encoded into text and visual tokens, which are then processed by transformer blocks to produce contextual embeddings shared with downstream experts through masked self-attention.
We further extend the multimodal inputs to tactile observations through a dedicated cross-attention mechanism, where tactile features serve as keys and values, and visual-language features act as queries. The resulting representations are then residually fused with the self-attention features, enabling joint reasoning over visual, linguistic, and tactile observations within a unified latent space.

\noindent\textbf{Generation Expert.}
The Generation Expert explicitly models visuo-tactile dynamics by predicting future visual and tactile observations from historical interactions, forming physically grounded multimodal world representations to guide downstream action generation. Despite the remarkable progress of recent video generation models in visual prediction, directly applying generative modeling to contact-rich manipulation remains challenging due to the strict real-time requirements of high-frequency robot control. Following \citet{cai2026internvla}, we adopt a parallel decoding strategy for future visuo-tactile generation. The proposed non-autoregressive paradigm significantly improves computational efficiency while remaining effective for modeling future visuo-tactile interactions.

\noindent\textbf{Action Expert.} 
Conditioned on the latent features from the understanding expert and generation expert, together with the robot proprioceptive states $q_t$, the action expert predicts an action chunk $a_{t:t+H}$ using a flow matching objective. Since arm and hand motions exhibit substantially distinct dynamics and control granularity, jointly modeling them through a shared denoising process may obscure fine-grained dexterous behaviors and weaken arm-hand coordination in high-DoF settings. To address this, we propose \textbf{Factorized Flow Matching}, which explicitly decouples arm and hand motion generation. Specifically, arm and hand actions are initialized from independent noise distributions and projected into separate token sequences through MLP layers. The tokens are jointly processed by transformer blocks for coordinated motion generation before being decoded into clean action trajectories. Both the generation expert and action expert adopt Qwen3~\cite{yang2025qwen3} as their backbone.

\begin{wrapfigure}{r}{0.3\linewidth}
    \vspace{-1.0em}
    \centering
    \includegraphics[width=\linewidth]{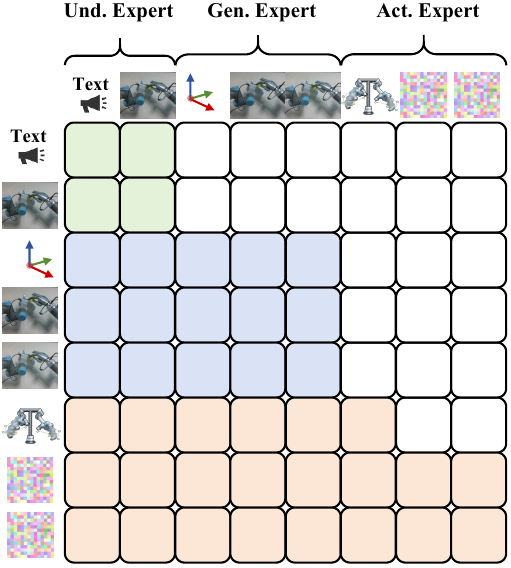}
    \caption{\textbf{Attention mask.}}
    \label{fig:attention_mask}
    \vspace{-1em}
\end{wrapfigure}
\noindent\textbf{Cross-Modal Joint Attention}. We implement a blockwise attention mask to control the information flow across experts, as shown in \Cref{fig:attention_mask}. Information is propagated unidirectionally from the understanding expert to the generation expert, and then to the action expert. Tokens in later experts can attend to tokens from preceding experts, while the reverse attention is prohibited. Within both the understanding expert and generation expert, tokens are bidirectionally attended for more effective multimodal reasoning and representation learning. In the action expert, state tokens can attend to themselves and all earlier blocks, while action tokens can attend to themselves, state tokens, and all preceding blocks. This asymmetric attention design enables action generation to fully condition on semantic understanding, imagined future interactions, and robot states.

\subsection{Adaptive Visuo-Tactile Fusion}
\label{Adaptive Visuo-Tactile Fusion}
\noindent\textbf{Tactile Feature Extraction.}
We use multi-finger deform maps as tactile inputs due to their rich geometric contact information. A shared ResNet-based~\cite{he2016deep} tactile encoder extracts tactile features from all fingertips. The extracted features are spatially pooled and projected into tactile tokens, which are subsequently fused through transformer layers to capture cross-finger interactions and global contact patterns. The tactile encoder finally produces two complementary representations: local tactile tokens that preserve fine-grained contact details for visuo-tactile interaction, and a global tactile token that summarizes the overall contact state for contact-aware tactile gating.

\noindent\textbf{Adaptive Visuo-Tactile Cross-Attention.} 
To incorporate tactile feedback into visual-language embeddings, we introduce a cross-attention mechanism at each transformer block. The local tactile tokens serve as keys and values, while the prefix projection of the transformer input provides the queries. The resulting fused output is then added to the embeddings from the block's joint self-attention via a residual connection, allowing the model to integrate tactile cues while preserving the original multimodal context.

\noindent\textbf{Contact-Aware Gating Mechanism.}
In dexterous manipulation, physical contact is often intermittent and only arises during specific phases of a task, rather than being continuously present. Directly injecting tactile signals into the policy at all times can introduce biases in action generation and perturb the distribution of visual attention. To address this, we introduce a Contact-Aware  Gating Mechanism that allows the policy to adaptively modulate the weights of tactile inputs based on the relevance of contact information at each timestep. 
Formally, the gated output is computed as:
\vspace{-0.2em}
\begin{equation}
\begin{gathered}
\tilde{X} = X + \sigma \odot \mathrm{CrossAttn}(Q_{vl}, K_{local}, V_{local}), \\
\sigma = \mathrm{Gate}(z_\mathrm{global}),
\vspace{-0.4em}
\end{gathered}
\end{equation}

where $X$ represents the original input embeddings to the transformer layer, $Q_{vl}$ is the query from visual-language embeddings, while $K_{local}$ and $V_{local}$ correspond to the local tactile tokens serving as keys and values. The gating weight $\sigma$ is derived from the global tactile token $z_\mathrm{global}$ through a lightweight projection function $\mathrm{Gate}(\cdot)$. This mechanism allows the model to selectively incorporate tactile feedback based on the overall contact state across all fingertips.

\subsection{Visuo-Tactile Latent Co-Imagination}
\label{Visuo-Tactile Latent Co-Imagination}

The generation expert performs future-oriented latent imagination to jointly model visuo-tactile dynamics. Given historical and current visuo-tactile observations, the model produces predictive latent representations that capture the underlying physical dynamics of the interaction.

\noindent\textbf{Tactile Latent Prediction.} 
6-DoF contact force provides a compact and physically meaningful representation of contact dynamics, capturing both force magnitude and torque information. For tactile latent prediction, the forces from each fingertip are first mapped into embeddings using a dedicated force encoder, which are then processed by the policy to produce a tactile latent. The resulting latent can subsequently be decoded into representations of different granularities, including raw images, deform maps, and 6-DoF force vectors. These representations capture tactile dynamics from multiple perspectives, supporting physically grounded scene understanding and action generation.

\noindent\textbf{Visual Latent Prediction.}  
Inspired by recent world action models~\cite{kim2026cosmos, feng2026harmowam, yuan2026fast}, we equip DeCAL with world knowledge by generating future visual latents from historical and current observations. This forward-looking representation provides temporally informed context that enhances planning and decision-making in contact-rich manipulation tasks. We encode the multi-view visual images using the Cosmos VAE tokenizer~\cite{agarwal2025cosmos}, which provides expressive visual latents while preserving fine-grained details. To efficiently handle long visual sequences, we further apply a token compression mechanism following~\citet{cai2026internvla}. Specifically, the $32\times32$ latent feature grid is downsampled to $4\times 4$ using a convolutional layer with an 8×8 kernel, reducing sequence length while retaining essential spatial information. The compressed tokens are then decoded in parallel to generate future visual frames, leveraging the KV cache from the understanding expert.

During training, the predicted visual latents are directly regressed toward the Cosmos VAE targets encoded from future visual observations. While tactile latents, conditioned on current 6-DoF forces, are indirectly supervised by reconstructing future raw tactile images, deform maps, and 6-DoF forces. At inference time, the generation expert jointly predicts future visual and tactile latents to provide forward-looking context for action generation, while the reconstruction decoders are omitted, avoiding unnecessary pixel-space reconstruction.


\section{Experiment}
\label{sec:Experiment}

In this section, we evaluate DeCAL through extensive experiments addressing two key questions: (1) How does DeCAL perform against state-of-the-art policies and generalize to OOD scenarios (\Cref{Results and Analysis}, \ref{Generalization}, \ref{Generation Results})? (2) How does each component contribute to the overall performance (\Cref{Ablation Studies})?


\subsection{Experiment Setup}
\label{Experiment Setup}
\noindent\textbf{Tasks.} We evaluate DeCAL on six contact-rich and fine-grained dexterous manipulation tasks, as illustrated in \Cref{fig:tasks}: (1) \textit{Wipe Vase}, (2) \textit{Erase Whiteboard}, (3) \textit{Assemble Parts}, (4) \textit{Twist Cap}, (5)\textit{ Pipetting}, (6) \textit{Screw Light Bulb}. Each task is collected with 100 high-quality demonstrations and evaluated with 20 trials by default. We collect expert demonstrations through a teleoperation system with MetaGlove Pro and VIVE Trackers, following \citet{fu2025metis}, where the gloves retarget human hand motion to dexterous hands and the trackers control the robotic arm end-effectors. More details about the teleoperation setup and task specifications are provided in the Appendix.

\noindent\textbf{Baselines and Evaluation Metrics.}
We compare DeCAL with two state-of-the-art VLA models, GR00T N1.6~\cite{bjorck2025gr00t} and InternVLA-A1~\cite{cai2026internvla}, as well as two tactile-based specialist policies, ViTacFormer~\cite{heng2025vitacformer} and DECO~\cite{li2026deco}, which are built upon different architectures. To enable a fairer comparison in terms of tactile modality usage, we further reproduce InternVLA-A1$^{t}$ by naively incorporating tactile deform maps and net-force signals into the VLM backbone. We use two metrics to evaluate model performance: Success Rate (SR), indicating the entire task is successfully completed, and Progress Rate (PSR), capturing the average completion ratio across all task stages.

\subsection{Results and Analysis}
\label{Results and Analysis}
\noindent\textbf{Results on Real-World Experiments.} As shown in \Cref{tab:Main_Experiments}, DeCAL achieves the highest success rate across all tasks, outperforming all baselines, while also attaining the best PSR on the majority of tasks. Vision-based policies perform reasonably well on simpler tasks, such as Wipe Vase and Pipetting. However, they struggle with contact-rich dexterous manipulation requiring precise physical interaction and fine-grained contact understanding. Policies equipped with tactile sensing can incorporate richer environmental feedback, improving physical interaction awareness. But naively injecting tactile signals may also disturb the distribution of visual representations, leading to reduced accuracy in visual grounding and grasping. This issue becomes particularly evident in the Assemble Parts task, where effective coordination between vision and tactile is crucial. Benefiting from the adaptive visuo-tactile fusion mechanism and the joint modeling of visuo-tactile dynamics, DeCAL enhances physical interaction awareness while preserving robust visual grounding capabilities, thereby leading to improved performance on fine-grained dexterous manipulation tasks.

\begin{table}
    \renewcommand{\arraystretch}{1}
    \centering
    \caption{\textbf{Main results of six real-world tasks.} Each experiment is evaluated with 20 trials. }
    \label{tab:Main_Experiments}
    \resizebox{\linewidth}{!}{
    \begin{tabular}{c cc cc cc cc cc cc}
        \toprule
        \multirow{2}{*}{\bf Method} &
        \multicolumn{2}{c}{\bf Wipe Vase} &
        \multicolumn{2}{c}{\bf Erase Whiteboard} &
        \multicolumn{2}{c}{\bf Assemble Parts} &
        \multicolumn{2}{c}{\bf Twist Cap} &
        \multicolumn{2}{c}{\bf Pipetting} &
        \multicolumn{2}{c}{\bf Screw Light Bulb} \\
        \cmidrule(lr){2-13}
        & \bf SR & \bf PSR & \bf SR & \bf PSR & \bf SR & \bf PSR & \bf SR & \bf PSR & \bf SR & \bf PSR & \bf SR & \bf PSR \\
        \midrule
        GR00T N1.6  & 30.0\% & 78.8\%  & 50.0\% & 55.0\% & 30.0\% & 60.0\% & 10.0\% & 46.7\% & \textbf{60.0\%} & 85.0\% & 10.0\% & 45.0\%  \\ 
        InternVLA-A1 & 65.0\% & 85.0\% & 50.0\% & 58.3\% & 15.0\% & 43.3\% & 5.0\% & 36.7\% & 35.0\% & 78.8\% & 30.0\% & \textbf{53.8\%}\\
        ViTacFormer & 80.0\% & 88.8\% & 45.0\% & 50.0\% & 15.0\% & 71.7\% & 65.0\% & 81.7\% & 55.0\% & \textbf{86.3\%} & 25.0\% & 46.3\%\\
        DECO & 90.0\% & 95.0\% & 60.0\% & 70.0\% & 35.0\% & 61.7\% & 70.0\% & 83.3\% & 45.0\% & 71.3\% & 35.0\% & 41.3\%\\
        InternVLA-A1$^t$ & 75.0\% & 87.5\% & 45.0\% & 51.7\% & 45.0\% & 65.0\% & 25.0\% & 55.0\% & 20.0\% & 58.8\% & 25.0\% & 31.3\% \\
        \rowcolor{gray!20} \bf DeCAL (Ours) & \textbf{100.0\%} & \textbf{100.0\%} & \textbf{80.0\%} & \textbf{86.7\%} & \textbf{65.0\%} & \textbf{85.0\%} & \textbf{80.0\%} & \textbf{93.3\% }& \textbf{60.0\%} & \textbf{86.3\%} & \textbf{40.0\%} & 48.8\% \\
        \bottomrule
    \end{tabular}
    }
\end{table}

\begin{figure}
    \centering

    \begin{minipage}[t]{0.47\linewidth}
        \centering
        \includegraphics[width=\linewidth]{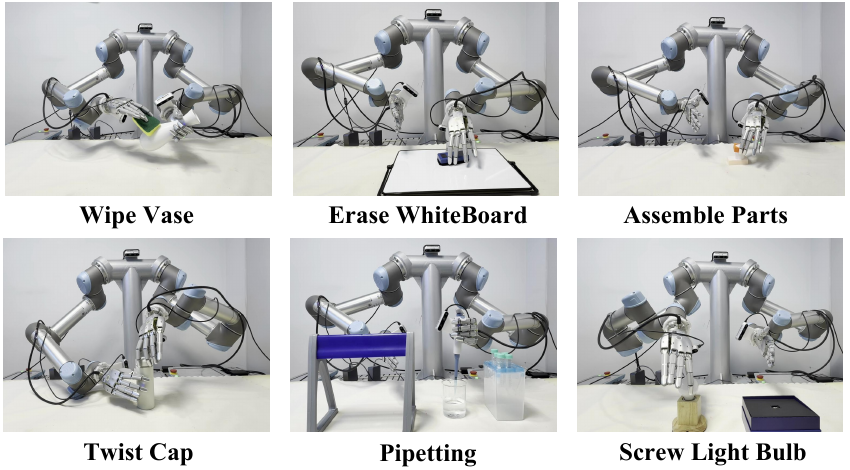}
        \vspace{-1.3em}
        \caption{\textbf{Visualization of dexterous tasks.}}
        \label{fig:tasks}
    \end{minipage}
    \hfill
    \begin{minipage}[t]{0.503\linewidth}
        \centering
        \includegraphics[width=\linewidth]{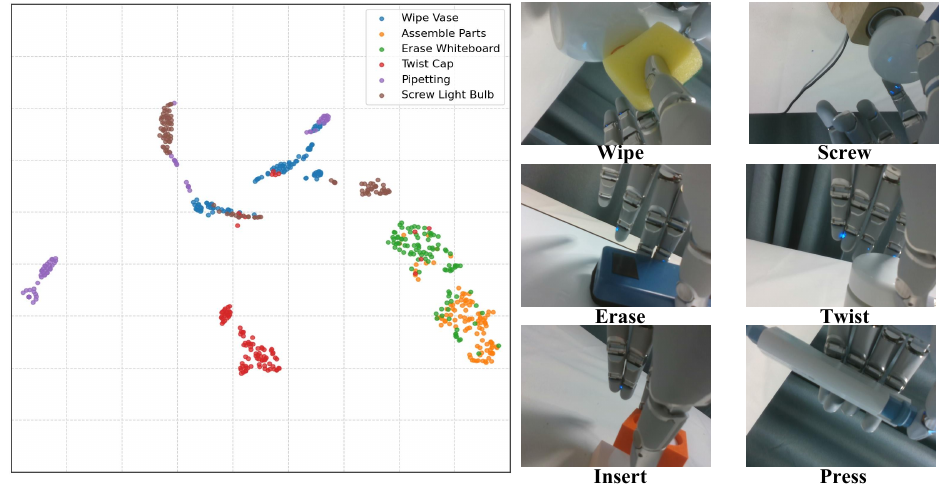}
        \vspace{-1.3em}
        \caption{\textbf{Visualize t-SNE of global tactile tokens.}}
        \label{fig:t-SNE_analysis}
    \end{minipage}
    \vspace{-2em}

\end{figure}

\noindent\textbf{Tactile Latent Analysis.} We analyze the global tokens extracted by the tactile encoder using t-SNE~\cite{van2008visualizing}. For each task, we sample frames from different contact phases, which correspond to diverse interaction patterns such as twisting, insertion, and wiping. As illustrated in \Cref{fig:t-SNE_analysis}, the learned global tactile representations exhibit clear clustering behavior under different contact modes, indicating that the encoder is able to capture meaningful and structured physical interaction patterns.

\begin{figure}
    \vspace{-1em}
    \centering
    \includegraphics[width=\textwidth]{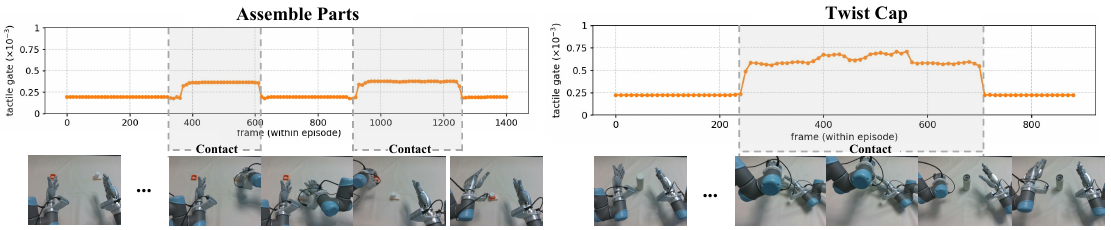}
    \caption{\textbf{Tactie gate.} The tactile gate value $\sigma$ increases during contact phases.}
    \label{fig:tactile_gate}
\end{figure}

\noindent\textbf{Adaptive Tactile Gate.} Next, we evaluate the effectiveness of the adaptive tactile gating mechanism. Taking the Assemble Parts and Twist Cap tasks as examples, \Cref{fig:tactile_gate} visualizes the tactile gate outputs across different frames within a testing episode. When the hand is not interacting with the environment, the tactile gate values remain consistently low, indicating that the policy is primarily dominated by visual signals. In contrast, once meaningful physical contact occurs between the fingers and the objects, the gate values increase significantly, allowing tactile information to play a more prominent role in the policy and influence action generation.

\begin{figure}
    \centering

    \begin{minipage}[t]{0.68\linewidth}
        \centering
        \includegraphics[width=\linewidth]{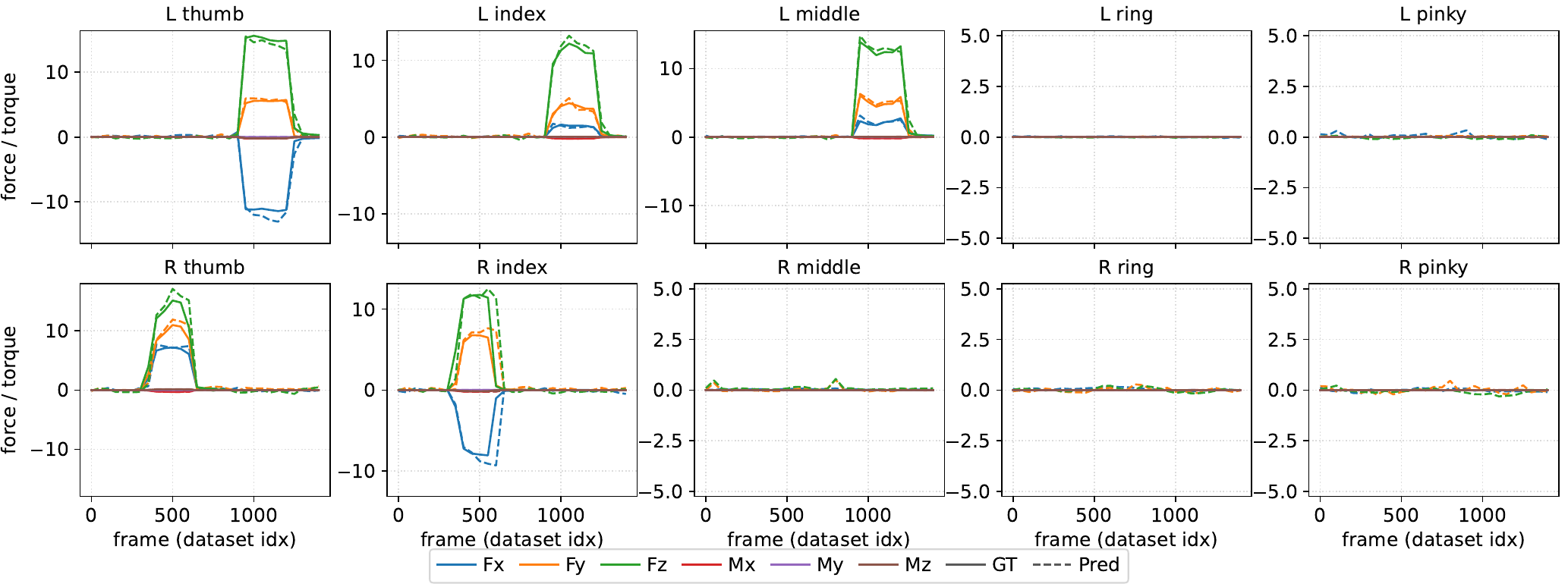}
        \caption{\textbf{Force prediction.}}
        \label{fig:force_prediction}
    \end{minipage}
    \hfill
    \begin{minipage}[t]{0.30\linewidth}
        \centering
        \includegraphics[width=\linewidth]{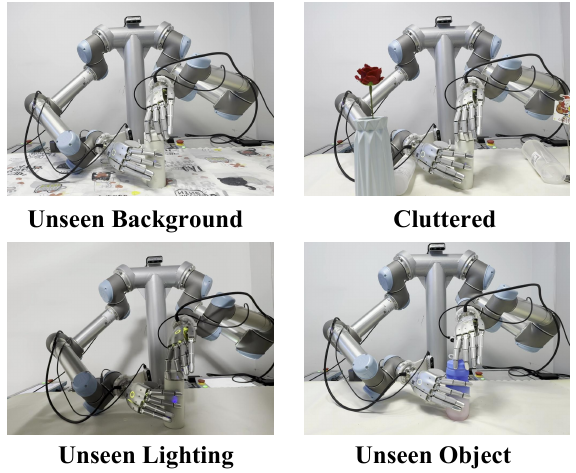}
        \caption{\textbf{OOD scenarios. }}
        \label{fig:generalization}
    \end{minipage}
    \vspace{-1em}

\end{figure}

\noindent\textbf{Tactile Force Prediction.}
We further compare the predicted tactile forces with the ground-truth values in the validation data. As shown in \Cref{fig:force_prediction}, the predicted 6D forces highly align with the ground-truth variations across different contact stages. This demonstrates that the proposed visuo-tactile latent co-imagination can effectively model underlying contact dynamics from multimodal observations. Moreover, the learned dynamics are further propagated to the action expert through joint attention, enabling more stable and precise action generation for contact-rich dexterous manipulation. 

\subsection{Generalization}
\label{Generalization}
Besides the remarkable effectiveness, DeCAL also demonstrates promising generalization capabilities under four \textit{out-of-distribution} scenarios: (1) \textbf{Unseen Background}, where a tablecloth perturbs the visual distribution. (2) \textbf{Cluttered Environment}, where random objects introduce visual distractions. (3) \textbf{Unseen Lighting}, where we use dimmer lighting conditions to simulate different illumination settings. (4) \textbf{Unseen Object}, where a novel cup with different shape, diameter, and height is introduced for manipulation, as shown in \Cref{fig:generalization}.
Taking Twist Cap as an example, we report the success rate of DeCAL, DECO, and ViTacFormer under four OOD scenarios in \Cref{fig:results generalization}. The results demonstrate that DeCAL generalizes effectively across various distribution shifts, achieving a 75.0\% success rate under the unseen object setting and substantially outperforming all baselines.

\subsection{Ablation Studies}
\label{Ablation Studies}

\begin{figure}
    \centering
    \begin{minipage}[t]{0.34\linewidth}
        \centering
        \includegraphics[width=\linewidth]{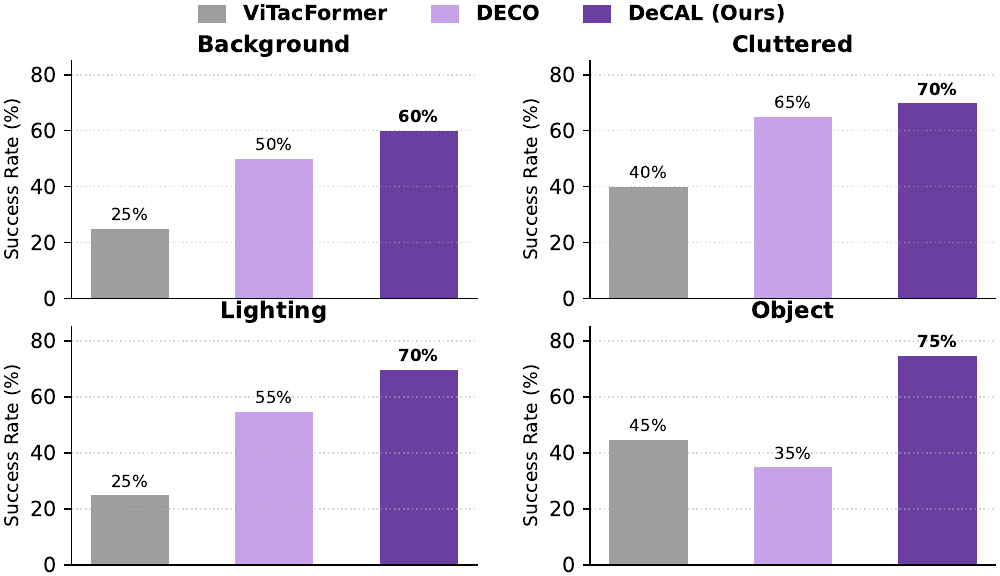}
        \vspace{-1.3em}
        \caption{\textbf{Generalization results.}}
        \label{fig:results generalization}
    \end{minipage}
    \hfill
    \begin{minipage}[t]{0.62\textwidth}
        \centering
        \vspace{-7.5em}
        \captionof{table}{\textbf{Ablation study of each component.}}
        \vspace{-0.4em}
        \label{tab:ablation}
        \scriptsize
        \setlength{\tabcolsep}{3pt}
        \begin{tabular}{c c c c | c c}
            \toprule
            \bf Factorized FM &
            \bf Tac Gating &
            \bf Vis Gen &
            \bf Tac Gen &
            \bf Assemble Parts &
            \bf Twist Cap \\
            \midrule
             \ding{55} & \ding{51} & \ding{51}
             & \ding{51} & 25.0\% & 35.0\%  \\
             \ding{51} & \ding{51} & \ding{55}
             & \ding{55} & 20.0\% & 30.0\% \\
             \ding{51} & \ding{51} & \ding{55}
             & \ding{51} & 35.0\% & 45.0\%\\
             \ding{51} & \ding{51} & \ding{51}
             & \ding{55} & 55.0\% & 60.0\%\\
             \ding{51} & \ding{55} & \ding{51}
             & \ding{51} & 50.0\% & 70.0\% \\
             \rowcolor{gray!20} \ding{51} & \ding{51} & \ding{51}
             & \ding{51} & 65.0\% & 80.0\%\\
            \bottomrule
        \end{tabular}
    \end{minipage}
    \vspace{-1em}
\end{figure}

In this section, we analyze the contribution of each component in DeCAL to the overall policy performance. We conduct ablation studies focusing on four aspects: (1) Factorized Flow Matching; (2) Tactile Gating Mechanism; (3) Visual Latent Generation; (4) Tactile Latent Generation. As shown in \Cref{tab:ablation}, Factorized Flow Matching contributes significantly to the overall task performance. Without this mechanism, we observe noticeable inconsistency and poor coordination between the arms and dexterous hands during manipulation. In addition, visual and tactile latent generation effectively models motion and contact dynamics, leading to more accurate action generation. Finally, the adaptive tactile gating mechanism enables the policy to dynamically adjust the importance of visual and tactile cues across different interaction stages, further improving the success rate on contact-rich manipulation tasks. More quantitative and qualitative results are provided in the Appendix.

\subsection{Generation Results}
\label{Generation Results}
\begin{figure}
    \centering
    \captionsetup{type=figure}
    \includegraphics[width=\textwidth]{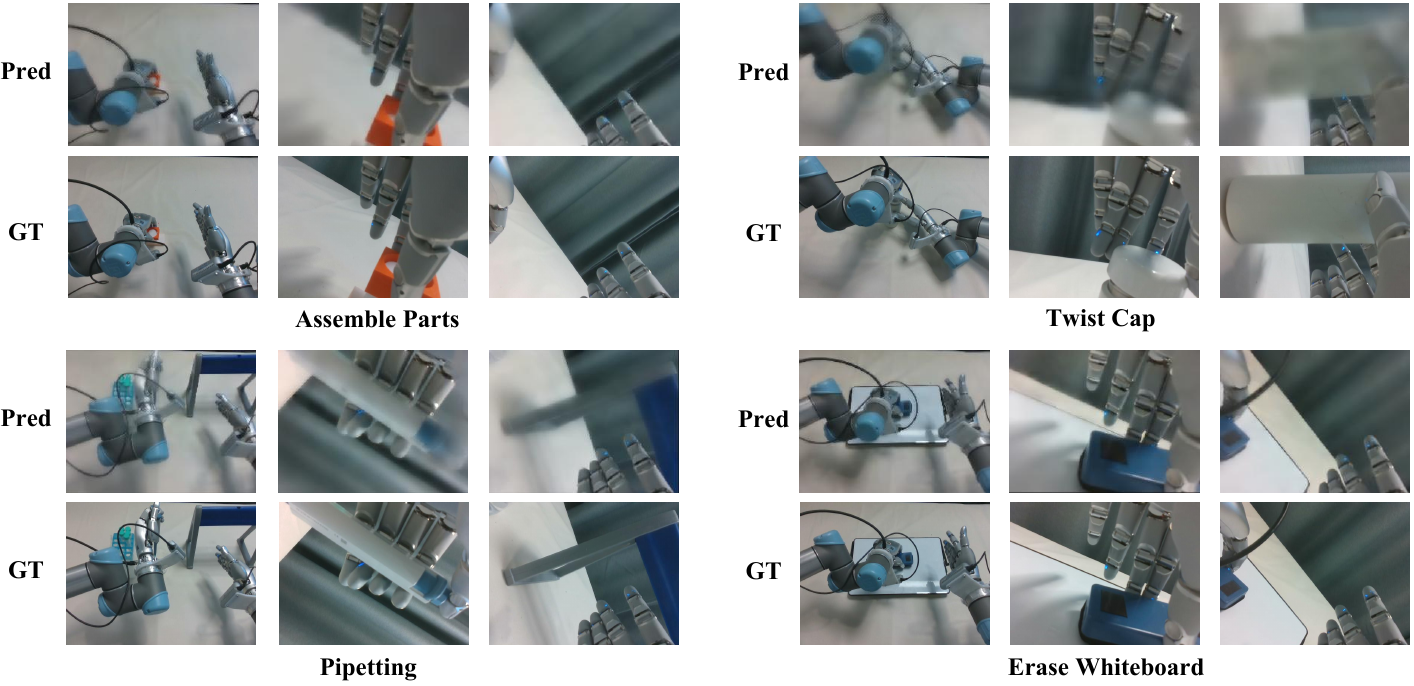}
    \caption{\textbf{Qualitative Results of Future Visual Generation.}}
    \label{fig:generation_results}
    \vspace{-1em}
\end{figure}

\begin{table}
    \vspace{-1em}
    \renewcommand{\arraystretch}{1}
    \centering
    \caption{\textbf{Visual Generation Results.}}
    \scriptsize
    \label{tab:generation}
    \begin{tabular}{c cc cc}
        \toprule
        \multirow{2}{*}{\bf Method} &
        \multicolumn{2}{c}{\bf Assemble Parts} &
        \multicolumn{2}{c}{\bf Twist Cap} \\
        \cmidrule(lr){2-5}
        & \bf Cos $\uparrow$ & \bf LPIPS $\downarrow$ & \bf Cos $\uparrow$ & \bf LPIPS $\downarrow$ \\
        \midrule
        InternVLA-A1 & 0.913 & 0.243 & 0.908 & 0.257\\
        \bf DeCAL (Ours) & \textbf{0.946} & \textbf{0.222} & \textbf{0.927} & \textbf{0.245}  \\
        \bottomrule
    \end{tabular}
    \vspace{-1em}
\end{table}
We evaluate future visual generation by comparing DeCAL with InternVLA-A1~\cite{cai2026internvla}, as both adopt a MoT architecture for future-frame prediction. Experiments are conducted on Assemble Parts and Twist Cap tasks, where models are trained on the training set and evaluated on the validation set by comparing the predicted frames against the ground-truth future observations. We assess generation quality using Cosmos feature similarity~(Cos $\uparrow$) and LPIPS~($\downarrow$), which measure semantic consistency and perceptual similarity, respectively. As shown in \Cref{tab:generation}, DeCAL consistently achieves better visual generation quality than InternVLA-A1. By incorporating tactile feedback, the model gains additional information about physical interactions and contact dynamics, enabling a more comprehensive understanding of environmental changes. Furthermore, the joint modeling of visuo-tactile dynamics helps the model better capture the evolution of future states under physical contact, resulting in more accurate and realistic future-frame predictions. More qualitative examples are presented in \Cref{fig:generation_results} and Appendix \ref{Supplementary Generation Results}.

\section{Conclusions}
\label{sec:conclusion}
In this paper, we present DeCAL, a physically-grounded dexterous vision-language-action model that effectively leverages tactile feedback for contact-rich manipulation. By unifying understanding, generation, and action within a collaborative framework, DeCAL enables coherent multimodal reasoning and fine-grained dexterous control. Furthermore, DeCAL adaptively fuses visuo-tactile information through contact-aware tactile gating and models visuo-tactile dynamics via future-oriented latent imagination, endowing the policy with implicit world knowledge. Extensive experiments on real-world dexterous tasks demonstrate the effectiveness and generalization of our approach.

\section{Limitations}
First, our method relies on accurate tactile perception, which can be affected by sensor noise, calibration errors, and model drift, potentially leading to performance degradation during long-term and high-load operation. Second, the teleoperation system does not provide fingertip force feedback to the operator, which may result in delayed contact adjustment, suboptimal contact regulation and limit the quality of tactile-aware demonstrations. Moreover, our current framework does not leverage large-scale visuo-tactile pretraining. We believe that scaling up diverse tactile dexterous data for pretraining could further improve the model’s capability, which we consider an important direction for future research.

\clearpage
\acknowledgments{This work was supported by the National Natural Science Foundation of China (62476011), the Beijing Natural Science Foundation (L252060). We would like to express our sincere gratitude to Yifan Ye and Yunfan Lou for their insightful discussions and valuable feedback on the methodological design. We are also grateful to Xiansheng Chen for his support with the hardware setup.}


\bibliography{example}  

\clearpage
 
\appendix
\input{appendix}

\end{document}

%% file: appendix.tex
\noindent{\large\bfseries Appendix}
\section{Additional Method Detail}
\subsection{\textbf{Training objective}}

DeCAL is trained with three complementary objectives: visual foresight generation, tactile foresight generation, and action prediction. For visual foresight, we supervise the predicted latent representations using the COSMOS latent features, encouraging the model to focus on task-relevant regions. For tactile foresight, the model predicts multiple forms of tactile representations (e.g., raw images, deformation maps, and 6-DoF force vectors), each supervised with a corresponding reconstruction loss to capture fine-grained and global contact dynamics. Action prediction is trained under the flow matching paradigm~\cite{lipman2022flow}, which models the transport velocity between noisy and target action distributions. The final training objective is a weighted sum of the three losses: $\mathcal{L}_\text{total} = \lambda_v \mathcal{L}_\text{visual} + \lambda_t \mathcal{L}_\text{tactile} + \mathcal{L}_\text{action}$.

DeCAL is trained with three complementary objectives: visual foresight generation, tactile foresight generation, and action prediction. For visual foresight, the model predicts future visual latent representations $\hat{z_v}$ conditioned on historical visual frames. These predictions are supervised using the pretrained COSMOS latent features $\hat{z_v}$, with a reconstruction loss that encourages the model to attend to task-relevant regions:
\begin{equation}
    \mathcal{L}_{visual} = \frac{1}{N_v}\sum_{i=1}^{N_v}||\hat{z_v}^{(i)} - z_v^{(i)}||_2^2,
\end{equation}
where $N_v$ denotes the number of visual tokens.

For tactile foresight, the model predicts multiple complementary future tactile representations: raw tactile images $\hat{R}$, deformation maps $\hat{M}$, and 6-DoF net force/torque vectors $\hat{F}$. Each is supervised with a corresponding reconstruction loss to capture both fine-grained local contact dynamics and global tactile interactions:

\begin{equation}
    L_{tactile} = \lambda_R \mathcal{L}_{img}(\hat{R}, R) + \lambda_M \mathcal{L}_{def}(\hat{M}, M) + \lambda_F \mathcal{L}_{force}(\hat{F}, F),
\end{equation}

where $\lambda_R$, $\lambda_M$, $\lambda_F$ balance the contributions of each tactile modality.

For action prediction, we adopt the flow matching paradigm~\cite{lipman2022flow}, which learns a transport vector field $v_\theta$ that maps a noisy action to the target action. Formally, let $s_t$ denote the proprioceptive state at time $t$, and let $\hat{a}_{t:t+k}^\tau$ be an interpolated noisy action chunk constructed as:
\begin{equation}
    \hat{a}_{t:t+k}^\tau = (1-\tau)\epsilon + \tau a_{t:t+k}, \quad \epsilon \sim \mathcal{N}(0, I), \quad \tau \sim Beta(1.5, 1.0),
\end{equation}

where $a_{t:t+k}$ is the expert action chunk, and $\tau$ progresses from 0 to 1 over K steps. Given the dataset of expert demonstrations $\mathcal{D}_2$, The model learns a velocity field $v_\theta(s_t, \hat{a}_{t:t+k}^\tau, h_{und}, h_{gen})$ that transports noisy actions toward the target, conditioned on contextual features from the understanding expert $h_{und}$ and generation expert $h_{gen}$:
\begin{equation}
    \mathcal{L}_{action} = \mathbb{E}_{\xi_2\sim \mathcal{D}_2}[||v_\theta(q_t, \hat{a}_{t:t+k}^\tau, h_{und}, h_{gen})-(a_{t:t+k}-\epsilon)||_2^2].
\end{equation}

The overall training objective is a weighted sum of the three losses:

\begin{equation}
    \mathcal{L}_{total} = \lambda_v \mathcal{L}_{visual} + \lambda_t \mathcal{L}_{tactile} + \mathcal{L}_{action}
\end{equation}

where $\lambda_v$ and $\lambda_t$ control the relative importance of visual, tactile objectives. This formulation encourages the model to jointly reason over multi-modal signals and generate physically grounded, contact-aware actions.

\subsection{Inference Pipeline}
During inference, DeCAL embeds multi-view visual observations, language instructions, and tactile inputs into prefix and middle tokens, which are processed by the understanding expert and generation expert. The prefix tokens produce contextual representations stored in KV caches, allowing subsequent computations to efficiently attend to all prior context. Subsequently, middle embeddings, including predicted visuo-tactile foresight, are integrated with these cached states to form latent representations that capture both task understanding and visuo-tactile dynamics. Finally, actions are generated iteratively via factorized flow matching, conditioned on the current state and the cached context. Running on a single NVIDIA RTX 4090 GPU, DeCAL achieves an average inference latency of 0.27 s per action chunk, demonstrating efficient inference for real-time dexterous manipulation.

\section{Policy Implementation Details}
For training DeCAL, we initialize the policy from Internvla-A1-3B~\cite{cai2026internvla} pretrained weights, which have been extensively trained on large-scale vision-language-action data and exhibit strong generalization across diverse tasks. Training is conducted on 8 NVIDIA H100 GPUs, with a per-device batch size of 4. We optimize the model using AdamW with the learning rate of $5.0e^{-5}$, along with a linear warmup (2,000 steps) and decay schedule down to $5.0e^{-6}$ over 100,000 steps. During training, the generation expert receives a temporal observation history consisting of the previous 15 frames (approximately 0.5 seconds of context at 30 Hz), together with the current observation, enabling the model to capture short-term visuo-tactile dynamics for action prediction. The policy predicts action chunks of length 50 during both training and inference.

\section{Real-Robot System}
\begin{figure}
    \centering
    \captionsetup{type=figure}
    \includegraphics[width=0.7\textwidth]{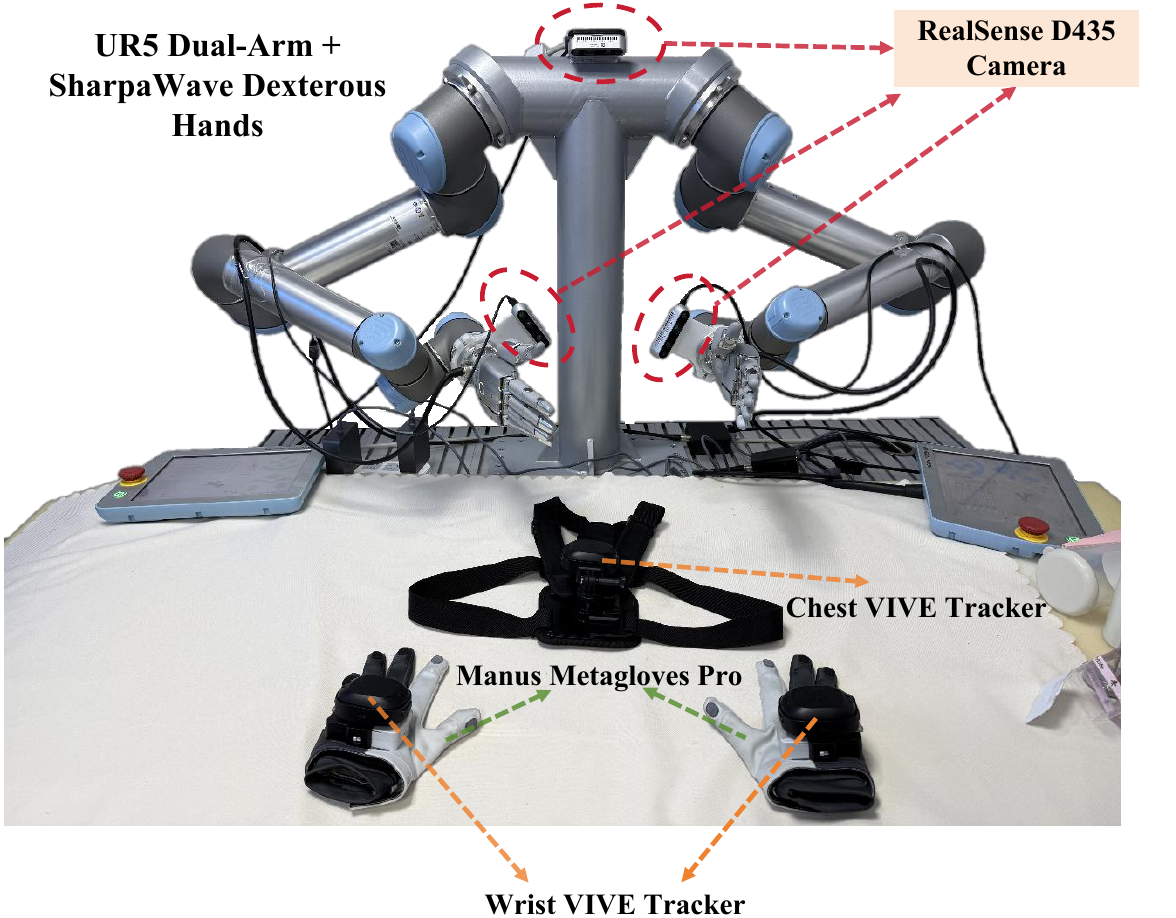}
    \caption{\textbf{Real Robot Platform.} Our setup consists of a dual UR5 robotic arm and a pair of SharpaWave five-finger dexterous hands. Visual observations are captured with one head-mounted and two wrist-mounted cameras, while high-resolution tactile data are collected from each fingertip. Expert demonstrations are acquired via a wearable glove-tracker teleoperation system.}
    \label{fig:robot_system}
    \vspace{-1em}
\end{figure}

As illustrated in \Cref{fig:robot_system}, our real-robot platform consists of a pair of 6-DoF UR5 robotic arms and two 22-DoF SharpaWave five-fingered dexterous hands. Visual observations are captured using two wrist-mounted cameras and one egocentric camera, all based on Intel RealSense D435 sensors, providing both local manipulation views and global scene perception. Each fingertip is further equipped with a high-resolution ($320 \times 240$) vision-based tactile sensor developed by Sharpa, enabling fine-grained perception of contact geometry and interaction dynamics during dexterous manipulation. The entire system operates at 30 Hz for real-time visuo-tactile control.

\textbf{Self-collected Robot Data}. We collect robot demonstrations through human teleoperation with a glove-tracker system~\cite{wang2024dexcap}. Three VIVE trackers are placed on the operator’s chest and wrists to record spatial positions of these key points. The wrist poses are then computed relative to the chest coordinate frame through coordinate transformation, and subsequently mapped to the arm joint configurations via inverse kinematics (IK) solvers. Simultaneously, Manus Metagloves Pro are employed to retarget human hand motions onto the dexterous hand, enabling accurate transfer of finger and joint movements. During demonstration, we capture visual observations from an egocentric view as well from the left and right wrist cameras. Each fingertip provides rich tactile signals, including three modalities: raw tactile images, deformation maps, and 6-DoF net forces. The entire system operates at 30Hz, balancing motion fidelity with reliable multi-sensor synchronization.

\section{Task Details}
\textbf{Wipe Vase.}
In this task, the robot must remove graffiti marks from a vase through coordinated bimanual manipulation. The robot first grasps and lifts the vase with its left hand, then picks up a sponge with its right hand. It continuously adjusts the vase orientation while maintaining stable contact between the sponge and the curved surface to remove the graffiti. After cleaning, the robot places both the vase and the sponge back onto the table. Success is achieved if all visible graffiti marks are removed and both objects are returned to their original locations.

\textbf{Erase Whiteboard.}
The robot is required to erase handwritten markings from a whiteboard. It first grasps an eraser with its left hand and establishes contact with the whiteboard surface. The robot then performs a controlled top-to-bottom wiping motion to remove the markings while maintaining stable contact throughout the cleaning process. Success is achieved if all visible markings are removed from the whiteboard.

\textbf{Assemble Parts.} The robot is required to assemble a plug-and-socket pair through precise dexterous manipulation. It first grasps and positions the white socket with its right hand, then picks up the orange plug with its left hand and aligns it above the insertion point. The robot must continuously adjust the plug pose based on contact feedback while performing the insertion. This contact-rich task requires accurate alignment and continuous adaptation to contact feedback during insertion. Success is achieved if the plug is fully inserted into the socket.

\textbf{Twist Cap.} 
The robot is required to unscrew a bottle cap through precise dexterous manipulation. It first stabilizes the bottle with its right hand, then grasps the cap using the thumb, index finger, and middle finger of its left hand. The robot gradually rotates the cap while maintaining a stable grasp until it is fully unscrewed, and subsequently places the cap on the table. Success is achieved if the cap is completely removed from the bottle and placed on the table.

\textbf{Pipetting.} The robot is required to dispense liquid using a pipette and subsequently return it to a holder. It first grasps the pipette with its left hand and moves it above a target beaker. The robot then presses the plunger with its thumb to release the liquid. After dispensing, the pipette is transferred to the right hand and placed onto a designated pipette holder. This task requires accurate thumb-actuated control, stable object handover, and precise placement. Success is achieved if the liquid is dispensed and the pipette is correctly returned to the holder.

\textbf{Screw Light Bulb.} The robot first grasps a light bulb from the workbench with its left hand and places it into the socket. It then uses the thumb and index finger of its right hand to rotate the bulb clockwise until it is fully tightened. This task requires coordinated finger movements and continuous contact feedback to maintain stable grasping and accuratlely regulate the screwing motion. success is achieved if the bulb is fully tightened in the socket and illuminates.

\begin{figure}[t]
    \centering
    \captionsetup{type=figure}
    \includegraphics[width=\textwidth]{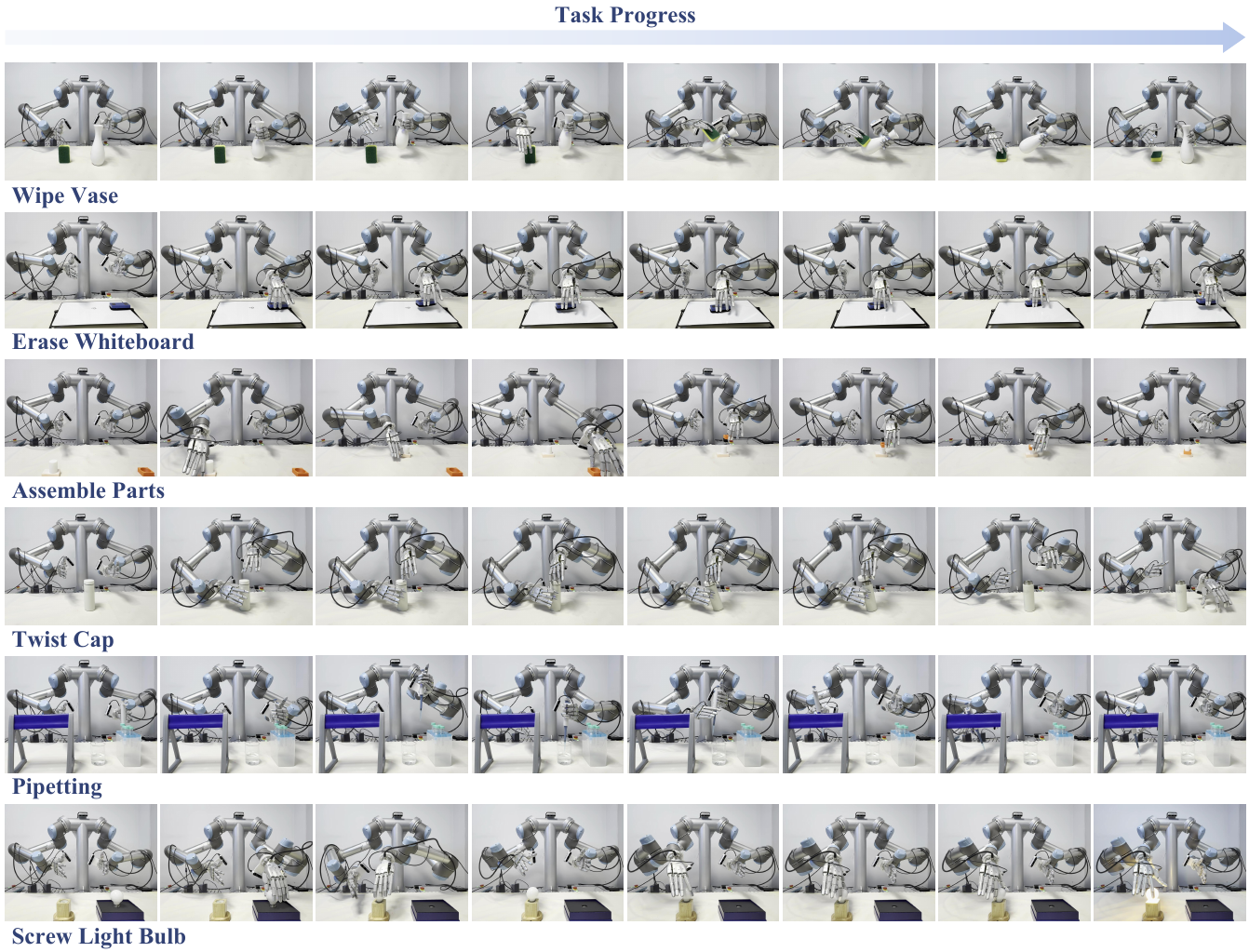}
    \caption{\textbf{Visualization of task progress.}}
    \label{fig:task_progress}
\end{figure}

The key manipulation steps for each task are summarized in \Cref{tab:subtasks}. Additional visualizations of the dexterous manipulation tasks can be found in \Cref{fig:task_progress}.

\begin{table}[t]
        \centering
        \captionof{table}{\textbf{Task descriptions}}
        \label{tab:subtasks}
        \begin{tabular}{c | c}
            \toprule
            \bf Task &
            \bf Key Manipulation Steps  \\
            \midrule
             Wipe Vase & \makecell[l]{\textbf{S1}: Grasp vase $\rightarrow$ \textbf{S2}: Grasp sponge $\rightarrow$\\ \textbf{S3}: Wipe vase surface $\rightarrow$ \textbf{S4}: Place vase}  \\
             Erase Whiteboard & \textbf{S1}: Pick up eraser $\rightarrow$ \textbf{S2}: Erase whiteboard $\rightarrow$ \textbf{S3}: Place eraser back \\
             Assemble Parts & \textbf{S1}: Pick place socket $\rightarrow$ \textbf{S2}: Pick plug $\rightarrow$ \textbf{S3}: Insert plug into socket\\
             Twist Cap & \textbf{S1}: Grasp bottle $\rightarrow$ \textbf{S2}: Unscrew cap $\rightarrow$ \textbf{S3}: Place cap \\
             Pipetting & \makecell[l]{\textbf{S1}: Grasp pipette $\rightarrow$ \textbf{S2}: Dispense liquid $\rightarrow$ \\
             \textbf{S3}: Hand over pipette → \textbf{S4}: Hang pipette} \\
             Screw Light Bulb & \makecell[l]{\textbf{S1}: Grasp bulb $\rightarrow$ \textbf{S2}: Insert bulb into socket $\rightarrow$ \\
             \textbf{S3}: Screw bulb $\rightarrow$ \textbf{S4}: Illuminate bulb} \\
            \bottomrule
        \end{tabular}
\end{table}

\section{Additional Quantitative and Qualitative Results}
\subsection{Supplementary Generation Results.}
\label{Supplementary Generation Results}

\begin{figure}
    \centering
    \captionsetup{type=figure}
    \includegraphics[width=\textwidth]{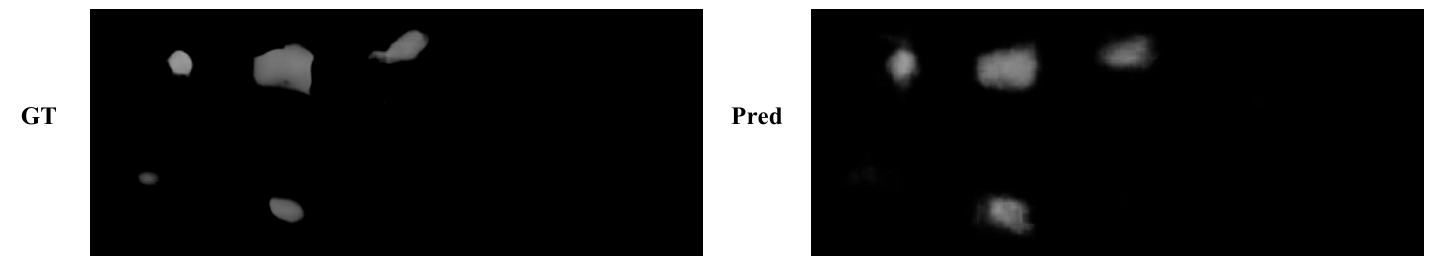}
    \caption{\textbf{Qualitative Results of Future tactile deform Generation.}}
    \label{fig:generation_tactile_results}
    \vspace{-1em}
\end{figure}

We further visualize the generated future tactile deformation maps in \Cref{fig:generation_tactile_results}. Although the generated tactile observations do not perfectly reconstruct fine-grained deformation details and exhibit some loss of local structure, they successfully preserve contact-related information. In particular, the high-response regions consistently correspond to fingertip-object contact areas, providing informative representations of future contact states. This suggests that the model is able to capture the underlying evolution of tactile interactions, which is sufficient for guiding downstream action generation.

\subsection{In-Domain (ID) Stage-Wise Results.}
\Cref{table: stage-wise id scenearios} reports the success rates of individual task stages under in-domain settings. Overall, DeCAL achieves the highest average stage success rate of 0.91, substantially outperforming all baselines. We observe that vision-based VLAs often perform well in early stages that primarily rely on visual grounding, such as object localization and grasping, but their performance degrades significantly in later stages involving sustained contact-rich interactions. Tactile specialist policies, such as ViTacFormer and DECO, generally achieve better performance during contact-intensive stages by leveraging direct physical feedback. However, the absence of explicit visuo-tactile dynamics modeling limits their understanding of physical interactions and future state evolution, leading to weaker long-horizon planning capabilities. In contrast, DeCAL consistently maintains strong performance across all stages by combining adaptive visuo-tactile fusion with explicit visuo-tactile dynamics modeling. For example, in Assemble Parts, DeCAL achieves substantially higher success rates in the final insertion stage, where precise alignment and contact-aware adjustment are essential. Similarly, in Twist Cap, DeCAL maintains a 100\% success rate during the cap-twisting stage, demonstrating its ability to reason about sustained physical interactions. 
\begin{table}
    \centering
    \scriptsize
    \setlength{\tabcolsep}{1.3pt}  
    \renewcommand{\arraystretch}{1.25}
    \caption{\textbf{Stage-wise success rates for ID scenarios}.}
    \label{table: stage-wise id scenearios}
    \begin{tabular}{c cccc ccc ccc ccc cccc cccc c}
    \toprule
    
    \multirow{2}{*}{\textbf{Method}}
    & \multicolumn{4}{c}{\textbf{Wipe Vase}}
    & \multicolumn{3}{c}{\textbf{Erase Whiteboard}}
    & \multicolumn{3}{c}{\textbf{Assemble Parts}}
    & \multicolumn{3}{c}{\textbf{Twist Cap}}
    & \multicolumn{4}{c}{\textbf{Pipetting}}
    & \multicolumn{4}{c}{\textbf{Screw Bulb}}
    & \multirow{2}{*}{\textbf{Avg}} \\
    \cmidrule(lr){2-5}
    \cmidrule(lr){6-8}
    \cmidrule(lr){9-11}
    \cmidrule(lr){12-14}
    \cmidrule(lr){15-18}
    \cmidrule(lr){19-22}
    & \multicolumn{4}{c}{S1$\rightarrow$S2$\rightarrow$S3$\rightarrow$S4}
    & \multicolumn{3}{c}{S1$\rightarrow$S2$\rightarrow$S3}
    & \multicolumn{3}{c}{S1$\rightarrow$S2$\rightarrow$S3}
    & \multicolumn{3}{c}{S1$\rightarrow$S2$\rightarrow$S3}
    & \multicolumn{4}{c}{S1$\rightarrow$S2$\rightarrow$S3$\rightarrow$S4} 
    & \multicolumn{4}{c}{S1$\rightarrow$S2$\rightarrow$S3$\rightarrow$S4} \\
    \midrule
    \textbf{GR00T N1.6} & \textbf{1.00} & 0.95 & 0.90 & 0.30 & \quad 0.60 & 0.55 & 0.50 & 0.80 & 0.70 & 0.30 & 0.85 & 0.45 & 0.10 & ~\textbf{1.00} &\textbf{ 1.00} & 0.80 & \textbf{0.60} & ~\textbf{0.90} & 0.60 & 0.20 & 0.10 & 0.69 \\
    \textbf{InternVLA-A1} & \textbf{1.00} & \textbf{1.00} & 0.75 & 0.65 & \quad 0.65 & 0.60 & 0.50 & 0.65 & 0.50 & 0.15 & 0.95 & 0.10 & 0.05 & ~\textbf{1.00} & \textbf{1.00} & 0.80 & 0.35 & ~0.75 & \textbf{0.70} & \textbf{0.40} & 0.30 & 0.68 \\
    \textbf{ViTacFormer} & \textbf{1.00} & 0.95 & 0.80 & 0.80 & \quad 0.55 & 0.50 & 0.45 & \textbf{1.00} & \textbf{1.00} & 0.15 & \textbf{1.00} & 0.80 & 0.65 & ~\textbf{1.00} & 0.95 & \textbf{0.95} & 0.55 & ~0.70 & 0.65 & 0.25 & 0.25 & 0.79 \\
    \textbf{DECO} & \textbf{1.00} & \textbf{1.00} & 0.90 & 0.90 & \quad0.75 & 0.75 & 0.60 & 0.80 & 0.70 & 0.35 & \textbf{1.00} & 0.80 & 0.70 & ~\textbf{1.00} & 0.75 & 0.65 & 0.45 & 0.45 & ~0.45 & 0.40 & 0.35 & 0.78 \\
    \textbf{InternVLA-A1$^t$} & 0.95 & 0.95 & 0.85 & 0.75 & \quad 0.60 & 0.50 & 0.45 & 0.80 & 0.70 & 0.45 & 0.90 & 0.50 & 0.25 & ~0.95 & 0.90 & 0.30 & 0.20 & ~0.35 & 0.35 & 0.30 & 0.25 & 0.64 \\
    \midrule
    \textbf{Ours} & \textbf{1.00} & \textbf{1.00} & \textbf{1.00} & \textbf{1.00} & \quad \textbf{0.95} & \textbf{0.85} & \textbf{0.80} & \textbf{1.00} & 0.90 & \textbf{0.65} & \textbf{1.00} & \textbf{1.00} & \textbf{0.80} & ~\textbf{1.00} & 0.95 & 0.90 & \textbf{0.60} & ~0.65 & 0.50 & \textbf{0.40} & \textbf{0.40} & \textbf{0.91}\\
    \bottomrule
    \end{tabular}
\end{table}

\subsection{Out-of-Domain (OOD) Stage-Wise Results}
We further report the stage-wise success rates on the Twist Cap task under four challenging generalization settings in \Cref{table: stage-wise ood scenearios}. While all methods experience performance degradation under distribution shifts, DeCAL exhibits better robustness across diverse scenarios, including changes in background, lighting conditions, scene clutter, and object geometry. ViTacFormer exhibits noticeably less stable hand motions under unseen environments, often resulting in object slippage during the cap-twisting stage. DECO demonstrates stronger robustness to environmental perturbations, but suffers a more pronounced performance degradation when manipulating unseen objects. In contrast, DeCAL consistently achieves superior performance across all generalization settings. Notably, under the \textit{Unseen Object} setting, where the test cups differ significantly in height, diameter, and shape, DeCAL still achieves a final-stage success rate of 0.75, exceeding the best baseline by 30\% points. We attribute this improvement to the joint modeling of visuo-tactile dynamics, which enables the policy to capture transferable interaction patterns beyond appearance-specific cues and therefore generalize more effectively to unseen scenarios.

\begin{table}
    \centering
    \scriptsize
    \setlength{\tabcolsep}{2.5pt}  
    \renewcommand{\arraystretch}{1.25}
    \caption{\textbf{Stage-wise success rates for OOD scenarios on Twist Cap.}}
    \label{table: stage-wise ood scenearios}
    \begin{tabular}{c ccc ccc ccc ccc c}
    \toprule
    
    \multirow{2}{*}{\textbf{Method}}
    & \multicolumn{3}{c}{\textbf{Unseen Background}}
    & \multicolumn{3}{c}{\textbf{Cluttered}}
    & \multicolumn{3}{c}{\textbf{Unseen Lighting}}
    & \multicolumn{3}{c}{\textbf{Unseen Object}}

    & \multirow{2}{*}{\textbf{Avg}} \\
    \cmidrule(lr){2-4}
    \cmidrule(lr){5-7}
    \cmidrule(lr){8-10}
    \cmidrule(lr){11-13}
    & \multicolumn{3}{c}{S1$\rightarrow$S2$\rightarrow$S3}
    & \multicolumn{3}{c}{S1$\rightarrow$S2$\rightarrow$S3}
    & \multicolumn{3}{c}{S1$\rightarrow$S2$\rightarrow$S3}
    & \multicolumn{3}{c}{S1$\rightarrow$S2$\rightarrow$S3}\\
    \midrule
    \textbf{ViTacFormer} & ~~0.80 & 0.60 & 0.25 & ~~\textbf{1.00} & \textbf{1.00} & 0.40 & ~~0.90 & 0.85 & 0.25 & ~~0.90 & \textbf{0.90} & 0.45 & ~0.69 \\
    \textbf{DECO} & ~~\textbf{0.90} & \textbf{0.75} & 0.50 & ~~\textbf{1.00} & 0.85 & 0.65 & ~~\textbf{0.95} & 0.60 & 0.55 & ~~0.70 & 0.60 & 0.35 & ~0.70 \\
    \midrule
    \textbf{Ours} & ~~0.75 & \textbf{0.75} & \textbf{0.60} & ~~\textbf{1.00} & 0.80 & \textbf{0.70} & ~~\textbf{0.95} & \textbf{0.90} & \textbf{0.65} & ~~\textbf{0.95} & \textbf{0.90} & \textbf{0.75} & ~\textbf{0.81} \\
    \bottomrule
    \end{tabular}
\end{table}

\subsection{Supplementary Generalization Analysis}
In the generalization experiments, we evaluate the models under four types of unseen scenarios, including unseen background, unseen lighting condition, cluttered scene, and unseen object. The first three settings mainly introduce environment-level perturbations, while the unseen-object setting introduces an object-level distribution shift. DECO and ViTacFormer benefit from random image augmentations during training, which improves their robustness to visual changes such as background variation, illumination shifts, and moderate scene clutter. 
\begin{wrapfigure}{r}{0.26\linewidth}
    \vspace{-1.0em}
    \centering
    \includegraphics[width=\linewidth]{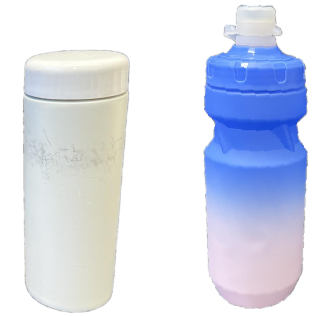}
    \caption{\textbf{Novel Object.}}
    \label{fig:unseen_object}
\end{wrapfigure}
However, the unseen-object setting is substantially more challenging. As shown in \Cref{fig:unseen_object}, we replace the training object with a novel cup that differs in color, diameter, shape, and height. For the Twist Cap task, such object-level changes require the policy to adjust not only the visual localization of the object, but also the grasping height, contact position, and wrist pose in real time. Both baselines show a clear performance drop under this object-level shift, suggesting that visual augmentation alone is insufficient for handling changes in contact dynamics. In contrast, DeCAL maintains a success rate of 75\%, demonstrating its stronger ability to generalize to the novel object by leveraging tactile feedback and future-oriented latent imagination.

\subsection{Supplementary Ablation Study Analysis}
We further provide a detailed analysis of the ablation results. (1) Removing Factorized Flow Matching leads to a noticeable performance drop. Without separating arm and hand motion during denoising, the policy models the full action space with a shared denoising process, which weakens arm-hand coordination. This often leads to inconsistent motions between arm-level reaching and fine-grained finger movements, especially in grasping and bimanual tasks where small pose or timing errors can cause unstable contact or failed execution. (2) Removing visual and tactile latent generation weakens performance in contact-sensitive tasks. Without this future-oriented visuo-tactile prediction, the policy struggles to anticipate how observations and contact states evolve after actions. In the Assemble Parts task, the policy may hesitate or oscillate near the socket during the ``insert plug into socket'' step, failing to make a consistent progress toward insertion. This indicates that latent generation provides useful predictive representations for modeling both motion and contact dynamics. (3) Removing the tactile gating mechanism reduces the policy's ability to adaptively balance visual-language features. Naively fusing tactile features with visual-language embeddings may introduce noise or interfere with visual grounding, especially when contact are weak. Introducing the tactile gating mechanism mitigates this issue by modulating the contribution of tactile features according to the current interaction state.

\section{Baseline Settings}
\textbf{For GR00T N1.6.~\cite{bjorck2025gr00t}} 
All experiments are conducted on a single NVIDIA H100 GPU. 
We use the full joint configuration of the dual-arm dual-hand system, with two 6-DoF robot arms and two 22-DoF dexterous hands, resulting in a 56-DoF action space for representing the complete joint action. 
The model is initialized from the pretrained GR00T-N1.6-3B checkpoint and fine-tuned for 50K steps with a batch size of 32. 
We optimize the model using AdamW with an initial learning rate of $1.0 \times 10^{-4}$. 
The learning rate is warmed up during the first $5\%$ of the training steps and then decayed using a cosine learning rate schedule.

\textbf{For InternVLA-A1~\cite{cai2026internvla}.} All experiments are conducted on 8 NVIDIA H100 GPUs. We use the full joint configuration of the dual-arm dual-hand system, consisting of two 6-DoF robot arms and two 22-DoF dexterous hands (a total of 56-DoF). We extend the maximum action dimensionality to support this full joint action representation and initialize the model from the pretrained InternVLA-A1-3B checkpoint, with newly introduced parameters randomly initialized. The model is trained with mixed precision for 100K steps, using a per-GPU batch size of 4, an action chunk size of 50, and AdamW with a learning rate of $5.0 \times 10^{-5}$. The learning rate is warmed up for the first 2K steps and then linearly decayed over 200K steps.

\textbf{For ViTacFormer~\cite{heng2025vitacformer}.} All experiments are conducted on a single NVIDIA H100 GPU. To adapt ViTacFormer to our visuo-tactile dexterous manipulation setting, we use the 6-D net force from each fingertip as the tactile input and the supervision signal, since the original model does not support tactile image inputs. We use ResNet-18 as the visual encoder and optimize it with a smaller learning rate of $1.0 \times 10^{-5}$, while the remaining modules, including the Transformer backbone, action prediction head, and tactile fusion layers, are randomly initialized and trained with a learning rate of $1.0 \times 10^{-4}$. We set the KL weight to 10, the action chunk size to 100, the hidden dimension to 512, and the batch size to 128. The model is trained for 2,000 epochs.

\textbf{For DECO~\cite{li2026deco}.} All experiments are conducted on 4 NVIDIA H100 GPUs. We adapt DECO to our visuo-tactile dexterous manipulation setting by using the 6-D net force from each fingertip as the tactile input. Following the original design, we adopt a pretrained ResNet-34 image encoder as the visual backbone and fine-tune it in a full-parameter manner. The model is optimized with SGD for 500 epochs, using a batch size of 1024. We set the initial learning rate to $1.0 \times 10^{-4}$ and use a cosine annealing schedule with a minimum learning rate of $5.0 \times 10^{-6}$. The first epoch is used for learning-rate warmup. 

\textbf{For InternVLA-A1$^t$.} For a fair comparison, we construct a tactile-augmented variant of InternVLA-A1, denoted as InternVLA-A1$^t$, by incorporating both tactile deforms and 6-D net force signals as additional VLM inputs, matching the input modalities used by DeCAL. Specifically, the current-frame deformation maps from 10 fingertips are encoded by a Multi-Finger Tactile Encoder with a shared CNN and an inter-finger Transformer, yielding compact tactile features. Meanwhile, each fingertip's 6-D force is projected into the hidden space by a lightweight MLP to obtain force features. These tactile and force features are concatenated and injected into each understanding Transformer block through an additional 8-head cross-attention layer, using the prefix hidden states as queries and the tactile features as keys and values, followed by a residual connection. All other hyperparameters, pretrained weights, camera inputs, and action settings are kept the same as InternVLA-A1.

\section{Failure Case Analysis}
Through extensive real-world experiments, we identify two primary categories of failure modes that can adversely affect the performance of DeCAL: The first category is related to \textbf{bimanual coordination}. For tasks that require precise synchronization between two arms, such as transferring a pipette from one hand to the other in the Pipetting task, small discrepancies in arm motion, hand alignment, or grasp timing may lead to unstable contact and eventually cause the object to slip or deviate from the desired pose. The second category arises from the \textbf{limited field of view} of the head-mounted camera. In tasks involving large-range motions, the manipulated object may occasionally move outside the visible region, making it difficult for the policy to maintain reliable visual tracking and state estimation. A potential direction for mitigating this limitation is to incorporate wider-angle or fisheye cameras, which can provide more comprehensive visual coverage of the workspace. Another promising direction is to introduce active perception, where the robot dynamically adjusts its viewpoint or selects informative observations during execution to better track task-relevant objects and reduce visual uncertainty.